\def\year{2022}\relax
\pdfoutput=1
\documentclass[letterpaper]{article} 
\usepackage{aaai22}  
\usepackage{amsfonts}
\usepackage{amsmath}
\usepackage{amsthm}
\usepackage{times}
\usepackage{booktabs}
\usepackage{multirow}
\newtheorem{theorem}{Theorem}[section]
\newtheorem{proposition}[theorem]{Proposition}
\usepackage{graphicx}
\usepackage{subfig}
\usepackage{helvet}
\usepackage{enumerate}
\usepackage{times}  
\usepackage{helvet}  
\usepackage{courier}  
\usepackage[hyphens]{url}  
\usepackage{graphicx} 
\usepackage{natbib}  
\usepackage{caption} 
\DeclareCaptionStyle{ruled}{labelfont=normalfont,labelsep=colon,strut=off} 
\usepackage{algorithm}
\usepackage{algorithmic}

\usepackage{newfloat}
\usepackage{listings}
\floatstyle{ruled}
\newfloat{listing}{tb}{lst}{}
\floatname{listing}{Listing}
\usepackage{lipsum}

\title{PetsGAN: Rethinking Priors for Single Image Generation}
\author {
    Zicheng Zhang\textsuperscript{\rm 1},
    Yinglu Liu\textsuperscript{\rm 2}, 
    Congying Han\textsuperscript{\rm 1}\thanks{Corresponding author},
    Hailin Shi\textsuperscript{\rm 2},
    Tiande Guo\textsuperscript{\rm 1},
    Bowen Zhou\textsuperscript{\rm 2}
}
\affiliations {
    \textsuperscript{\rm 1} University of Chinese Academy of Sciences \\
    \textsuperscript{\rm 2} JD AI Research\\ zhangzicheng19@mails.ucas.ac.cn, liuyinglu1@jd.com, hancy@ucas.ac.cn \\
    shihailin@jd.com, tdguo@ucas.ac.cn, bowen.zhou@jd.com
}
\usepackage{bibentry}

\begin{document}

\maketitle
\let\thefootnote\relax\footnotetext{This work was conducted when Zicheng Zhang was doing internship at JD AI Research.}
\let\thefootnote\relax\footnotetext{\textsuperscript{\rm 1}Code: \url{https://github.com/zhangzc21/PetsGAN}}

\begin{abstract}
Single image generation (SIG), described as generating diverse samples that have similar visual content with the given single image, is first introduced by SinGAN which builds a pyramid of GANs to progressively learn the internal patch distribution of the single image. It also shows great potentials in a wide range of image manipulation tasks.
However, the paradigm of SinGAN has limitations in terms of generation quality and training time. Firstly, due to the lack of high-level information, SinGAN cannot handle the object images well as it does on the scene and texture images. Secondly, the separate progressive training scheme is time-consuming and easy to cause artifact accumulation.
To tackle these problems, in this paper, we dig into the SIG problem and improve SinGAN by fully-utilization of internal and external priors. The main contributions of this paper include:
1) We introduce to SIG a regularized latent variable model. To the best of our knowledge, it is the first time to give a clear formulation and optimization goal of SIG, and all the existing methods for SIG can be regarded as special cases of this model.
2) We design a novel Prior-based end-to-end training GAN (PetsGAN)  to overcome the problems of SinGAN. For one thing, we inject external priors obtained by the GAN inversion with PetsGAN to alleviate the problem of lack of high-level information for generating natural, reasonable and diverse samples, even for the object images. For another, we inject the internal priors  of patch distribution correction with PetsGAN to ease and speed up the patch distribution learning. Moreover, our method gets rid of the time-consuming progressive training scheme and can be trained end-to-end. 3) We construct abundant qualitative and quantitative experiments to show the superiority of our method on both generated image quality, diversity, and the training speed. Moreover, we apply our method to other image manipulation tasks (\textit{e.g.}, style transfer, harmonization), and the results further prove the effectiveness and efficiency of our method\textsuperscript{\rm 1}. 

\end{abstract}

\begin{figure}[t]
	\centering
	\includegraphics[width=0.98\linewidth]{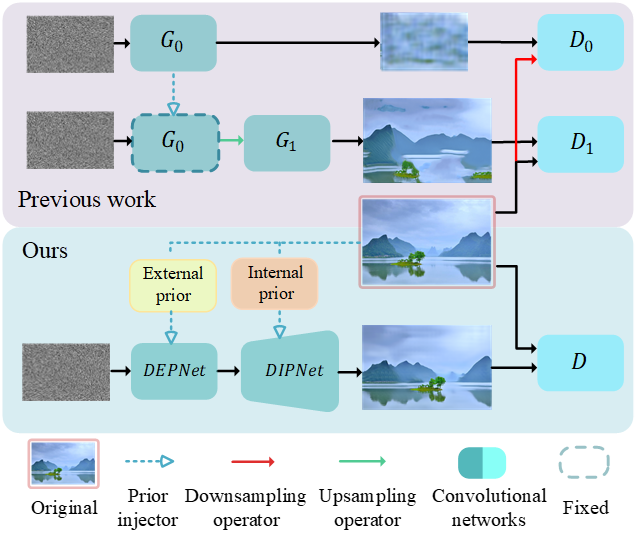}
	\caption{Comparing PetsGAN with previous work. The upper part shows the paradigm of two-stage SinGAN, where the generator $G_0$ is trained on low resolution images first, then copied to the next stage with fixed weights for high resolution generation. 
	In contrast, PetsGAN is trained in one-stage by making full use of external and internal priors, which embodies significant advantages in both training efficiency and generated image quality and diversity. See more details in Sec. \ref{PetsGAN}.}
	\label{fig:petsgan}
\end{figure}

  \begin{figure*}[htpb]
    \centering
    \includegraphics[scale=0.67]{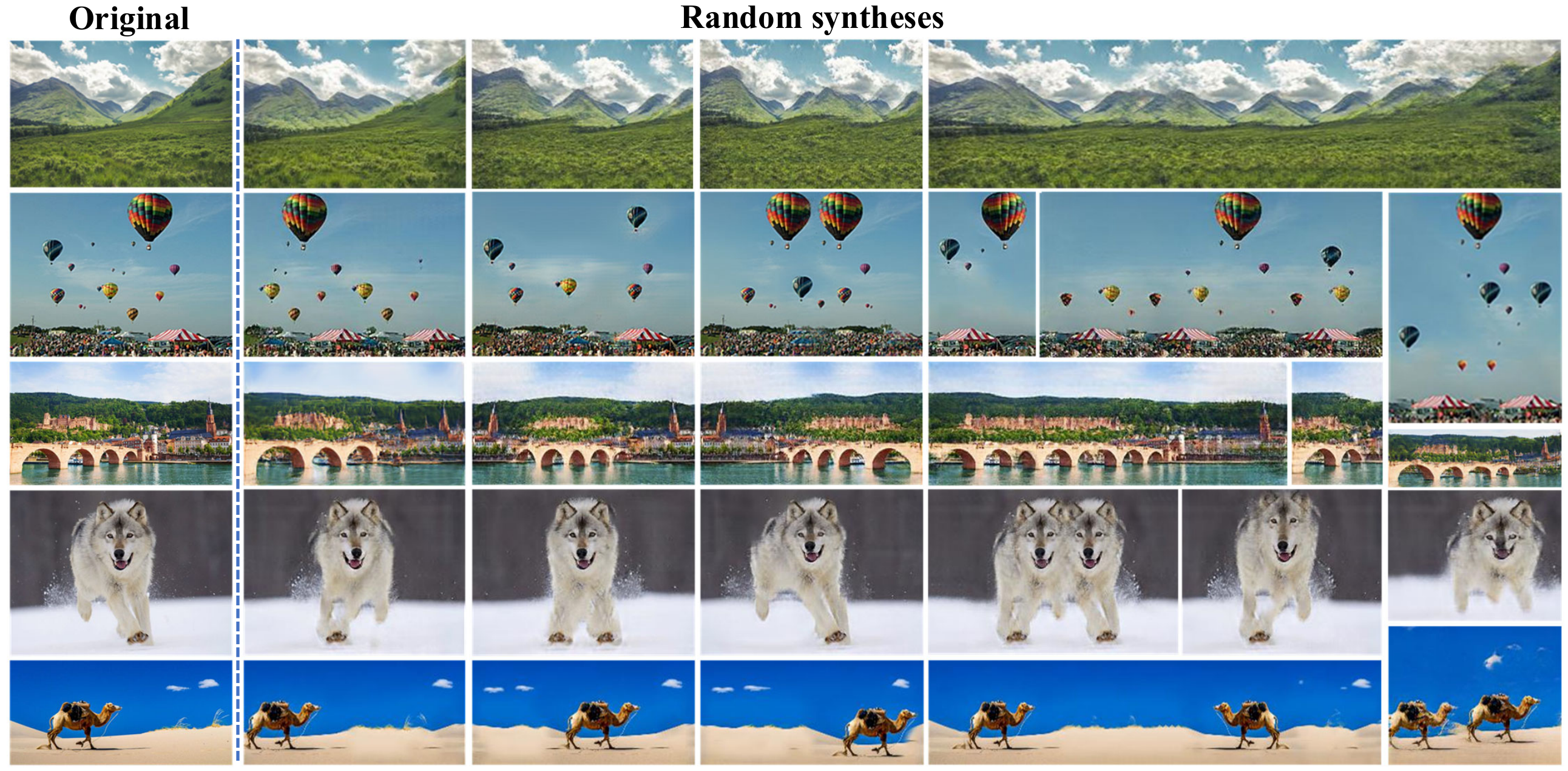}
    \caption{The first column refers to the original images, and the rest are random syntheses by PetsGAN. We can observe that PetsGAN can synthesize samples with arbitrary sizes, reasonable structure, fine texture, and great visual diversity.}
    \label{fig:show}
\end{figure*}
\section{Introduction}
The ability of modern deep learning to fulfill computer vision tasks is closely related to the building of large training sets \cite{deng2009imagenet,yu2015lsun}. 
It is, however, costly and difficult to collect training data in some scenarios, \textit{e.g.}, medical diagnosis, industrial defects. Generative models, such as generative adversarial networks (GANs) \cite{goodfellow2014generative} and variational autoencoder (VAE) \cite{kingma2013auto}, have been expected to augment the training dataset, but all previous and current state-of-the-art models like DCGAN \cite{radford2015unsupervised}, BigGAN \cite{brock2018large} and StyleGAN \cite{karras2019style,Karras2020AnalyzingAI}, face the same chicken-and-egg problem, which means that training generative models also requires a lot of data.  As a new research direction that fully exploits the internal statistics of natural signals by deep learning, deep internal learning \cite{ulyanov2017improved,shocher2018zero} has taken off and attracted wide attention since it only requires test image itself without additional supervision. 

  Not long ago deep internal learning has surprisingly stepped into the task of single image generation. Different from and beyond single image texture generative models \cite{jetchev2016texture,zhou2018non} that synthesize images with repeated texture patterns, single image generation models (SIG models) \cite{Shaham2019SinGANLA,Shocher2019InGANCA} are learned from the natural image and generate \textit{similar} but \textit{different} samples from the given image, where the former means similar visual content and the latter means diverse layout and appearance. The pioneering SinGAN \cite{Shaham2019SinGANLA} is the first unconditional SIG model, which builds a Laplacian pyramid of GANs with fully convolutional generators and patch discriminators to learn different scale information within the single image. The impressive performance of SinGAN proves the feasibility of internal learning on the generation task. From then on, many SIG models \cite{Hinz2020ImprovedTF,Chen2021MOGANMG,Zhang2021ExSinGANLA,Sushko2021OneShotGL,Granot2021DropTG} are proposed to explore along this line.


\par Although great breakthroughs and achievements have been made in this field, there are still many problems to be solved. 
Firstly, due to only acquiring the internal texture information of the given image, previous SIG models usually generate chaotic structures, which is especially serious on object images because the semantics and structural relationship between object parts are very important to synthesize plausible images. To alleviate this problem, \cite{Wang2021IMAGINEIS} utilize a pre-trained classification model to provide semantic information for image generation. However, the supervision is too coarse to guide high-fidelity image generation, and the model generalization is limited because 1) each generated sample should be optimized separately and 2) different classification models are needed for different image types.
Secondly, almost all SIG models \cite{Shaham2019SinGANLA,Hinz2020ImprovedTF,Chen2021MOGANMG,Zhang2021ExSinGANLA}  follow the paradigm of SinGAN \cite{Shaham2019SinGANLA} to equip the pyramid of generators and patch discriminators for internal patch learning. Although the progressive or cascaded learning of pyramid networks \cite{Denton2015DeepGI,karras2017progressive} reduces the difficulties of model optimization, the training is time-consuming and easy to cause artifacts accumulation. 

\par In this paper, we concentrate on the above-mentioned prominent problems in the SIG task, and make contributions as follows:
\par 1) We dive into the SIG problem in essence, that is, to learn a diverse distribution from the Dirac distribution composed of a single image. Obviously, it is an ill-posed restoration problem, and appropriate priors and regularization terms are the keys to solve it. Therefore, we construct a unified regularized latent variable model to formulate the SIG task. To the best of our knowledge, it is the first time to give a clear mathematical definition and optimization goal of this task. All the existing SIG models can be regarded as special cases of this model.
\par 2) We design a novel Prior-based end-to-end training single image GAN (PetsGAN), which is infused with internal priors and external priors to overcome the problems of SinGAN and the derived methods (Fig.~\ref{fig:petsgan}). For one thing, we inject external priors obtained by the GAN inversion with a DEPNet to alleviate the problem of lacking high-level information for generating natural, reasonable and diverse samples, even for the object images. For another, we inject the internal priors  of patch distribution correction with a DIPNet to ease and speed up the patch distribution learning. Benefiting from the fully-utilization of external and internal priors, our method gets rid of the time-consuming progressive training scheme and can be trained end-to-end. 
\par 3) We construct abundant qualitative and quantitative experiments on a variety of datasets. Fig. \ref{fig:show} shows some syntheses of PetsGAN,  experimental comparisons results show our method significantly surpasses other SIG models on image quality, diversity, and training speed. Moreover, we apply our method to other applications (\textit{e.g.}, fast training of high resolution image generation, style transfer, harmonization). The results further prove the effectiveness and efficiency of our method.

\section{Related Work}
\paragraph{Single image generation model.} 
Begin with modeling the internal patch distribution of a single image, SIG models aim to generate new plausible images with high texture quality. 
InGAN \cite{Shocher2019InGANCA} first proposes the conditional SIG model with the encoder-decoder architecture to make the geometric transformation of image look more natural.  SinGAN ~\cite{Shaham2019SinGANLA}  first proposes the unconditional SIG model composed of a Laplacian pyramid of GANs with fully convolutional generators and patch discriminators to learn different scale information. ConSinGAN ~\cite{Hinz2020ImprovedTF} modifies the training strategy of PGGAN \cite{karras2017progressive} to improve SinGAN by training multiple stages simultaneously. PatchGenCN \cite{Zheng_2021_CVPR}  proposes a pyramid of energy functions to explicit express the distribution within a single natural image. GPNN ~\cite{Granot2021DropTG} replaces each stage of SinGAN with a non-training patch match module  ~\cite{Barnes2009PatchMatchAR} to speed up the generation. At the same time, a lot of work is no longer limited to simple texture generation. MOGAN \cite{Chen2021MOGANMG} follows SinGAN but synthesizes the hand-marked regions of interest and the rest of the image separately, and then merges them into an unbroken image. One-shot GAN \cite{Sushko2021OneShotGL} is an end-to-end model with multiple discriminators for learning different features of image,  but it is not fully convolutional and more like normal GANs with diversity regularization \cite{Yang2019DiversitySensitiveCG}.    ExSinGAN  ~\cite{Zhang2021ExSinGANLA} introduces GAN inversion \cite{pan2020exploiting} and perceptual loss \cite{Johnson2016Perceptual} into SinGAN to improve performance on non-texture images. 

\paragraph{Deep priors.} In image restoration, priors mean some empirical knowledge or objective laws, which characterize the properties of the image and restrict the solution space. It can be explicitly expressed as a regularization term, or implicitly modeled in the loss function. For example, natural images are usually smooth in a local region \cite{Buades2005ANA,Mei2008VideoCP}, thus the smoothness can be taken as a prior in restoration tasks by total variation regularization \cite{Rudin1992NonlinearTV}. Deep image prior \cite{ulyanov2018deep} demonstrates that the randomly initialized deep convolutional network can be miraculously considered as an implicit regularization term, which performs much better than the traditional regularization terms in restoration tasks. From then on, various deep priors \cite{Gandelsman2019DoubleDIPUI,gu2020image,Wang2021TowardsRB} have been widely studied and applied in various fields. Deep generative prior (DGP) \cite{pan2020exploiting} assumes that the given image is sampled from the distribution space of the pre-trained generator, thus the generator is taken as an implicit regularization terms to restore image. IMAGINEA~\cite{Wang2021IMAGINEIS} adopts the similar idea that the pre-trained classifier is utilized as the implicit regularization terms to exploit the semantics knowledge of a given image.

\begin{figure}
	\centering
	\includegraphics[scale=0.4]{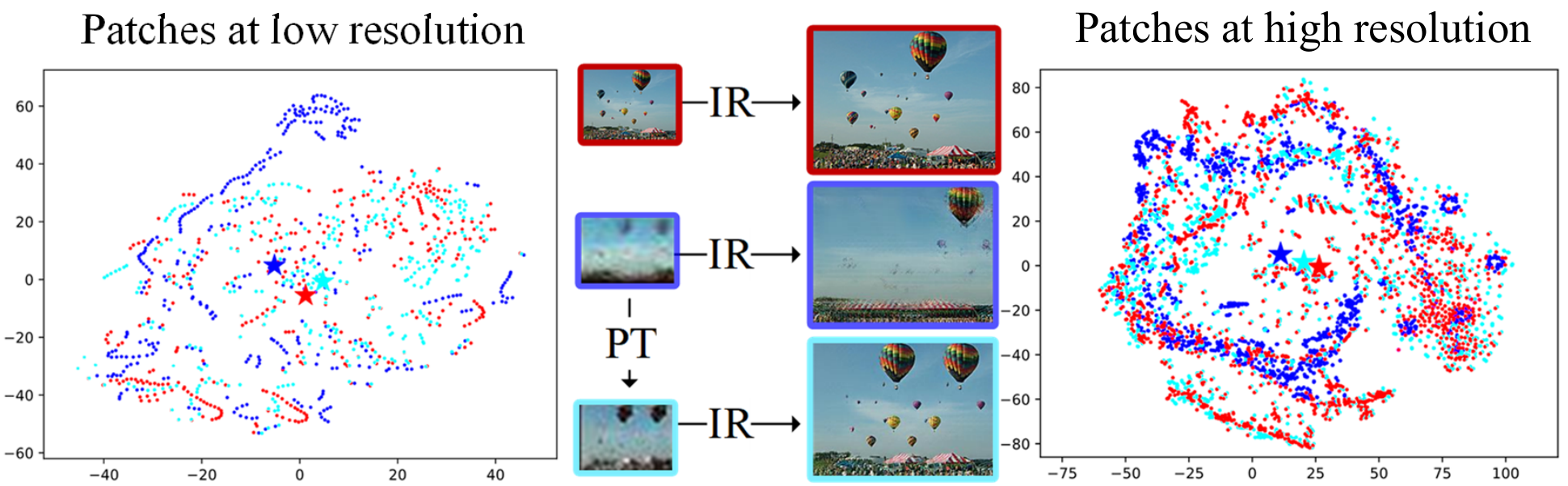}
	\caption{
	IR: image restoration network.
	PT: patch transfer.
	The scatters are the patches of low (left) and high (right) resolution, visualized by t-SNE.
	Red, blue and cyan represent the original, synthesis from DEPNet, and patch-transfer synthesis. 
	The ``star'' represents the mean value.  
	The distribution of cyan data can cover red data well, while a conspicuous deviation (caused by foreground patches) can be found between the blue data and red data.
	}
	\label{fig:patch_dis}
\end{figure}

\section{Methodology}

\subsection{Preliminaries} 
\paragraph{Notations.} To simplify the expression, we assume that all data of $d$ dimension comes from the set $\mathcal{X}^d$, and a measure space $(\mathcal{X}^d, \Sigma^d, \mu)$ can be defined over $\mathcal{X}^d$, where $\Sigma^d$ is a Sigma algebra and $ \mu$ is a measure. ${\rm Prob}(\mathcal{X}^d)$ is the space of probability measures defined on $\mathcal{X}^d$. $\mathbb{P}_{real} \in {\rm{Prob}}(\mathcal{X}^d)$ represents the real distribution of data, which is form-agnostic but there exists a sample set $\mathcal{D}=\{{\boldsymbol{x_i}} \in \mathcal{X}^d\}_{i=1}^n $. $\delta_{\boldsymbol{x}}$ represents the Dirac measure/distribution concentrated at ${\boldsymbol{x}}$.
$\rho$ denotes a generalized metric measuring the statistical distance between distributions.  $\operatorname{supp}(\mathbb{P}):=\left\{\boldsymbol{x} \in \mathcal{X}^d \mid \forall N_{\boldsymbol{x}} \in \Sigma^d, \mathbb{P} \left(N_{\boldsymbol{x}}\right)>0\right\}$ denotes the support of distribution $\mathbb{P}$, where $N_{\boldsymbol{x}}$ is the neighborhood of $\boldsymbol{x}$. 

\paragraph{Generalized Generation problem.} 
Generation task aims at learning a parameterized distribution  $\mathbb{P}_{\theta}$ to approximate the real distribution $\mathbb{P}_{real}$ through the empirical distribution ${\mathbb{P}}_{\mathcal{D}}$,
\begin{equation} \label{generalized loss}
    \min_{\theta} \rho({\mathbb{P}}_{\mathcal{D}},{\mathbb{P}}_{\theta}).
\end{equation}
There are extensive studies on problem (\ref{generalized loss}), \textit{e.g.}, the explicit density  models  \cite{kingma2013auto,Rezende2015VariationalIW,Oord2016PixelRN}  based on maximum log-likelihood estimation in which $\rho$ is formed by KullbackLeibler (KL) divergence, and implicit density models like GANs in which $\rho$ are formed by Jensen-Shannon (JS) divergence \cite{goodfellow2014generative}, Wasserstein distance \cite{arjovsky2017wasserstein}, \textit{etc.}
 
 \paragraph{Patch Distribution.}
 Suppose that $1\leq j,s \leq  d$ represent index and size of sliding window, respectively. For $\forall \  \boldsymbol{x}=[x_1,\dots,x_d]^{T} \in \mathcal{X}$,  its $j $-th patch $\boldsymbol{\omega}^{\boldsymbol{x},s}_j = [x_{ {\rm{mod}}(j-\lfloor s/2 \rfloor,d)}, \dots, x_{{\rm{mod}}(j+\lfloor s/2 \rfloor,d)}]^{T}$. Denoting the mapping $\pi: \Sigma^d \rightarrow \Sigma^s, \mathcal{D} \rightarrow  {\mathcal{D}}^{s} :=\mathop{\cup}\limits_{\boldsymbol{x}\in\mathcal{D}}\{\boldsymbol{\omega}^{\boldsymbol{x},s}_j\}_{j=1}^{d} $. The random variable ${\boldsymbol{\omega}}^{s}$ defined on dataset ${\mathcal{D}}^{s}$ also has the empirical distribution $\mathbb{P}^s_{\mathcal{D}}$, which is called the \textit{patch distribution} of $\mathcal{D}$ with size $s$. 
 Furthermore, $\pi$ induces the mapping $\tau: {\rm{Prob}}(\mathcal{X}^d) \rightarrow {\rm{Prob}} (\mathcal{X}^s)$, $\mathbb{P}_{\mathcal{D}} \rightarrow \mathbb{P}^s_{\mathcal{D}}$, and $\rho_{\tau}
 :=\rho(\tau(\cdot),\tau(\cdot))$ is the induced metric for patch distribution.
 
\subsection{Single Image Generation Problem}
The SIG problem is first roughly described by \cite{Shaham2019SinGANLA} as generating diverse samples that have the same visual content as the given natural image. However, there is no clear mathematical definition so far. To better interpret and solve this problem, we introduce a regularized latent variable model and all the existing methods for SIG can be regarded as special cases of this model.
\paragraph{Problem statement.} 
Given an image $\mathbf{I}$, the SIG problem is to learn a non-degenerate and realistic distribution $\mathbb{P}_{\theta}$ to restore the real distribution from the empirical distribution ${\mathbb{P}}_{\mathcal{D}}$. It's worth noting that ${\mathbb{P}}_{\mathcal{D}}$ degenerates into a single-point distribution ${\delta}_{\textbf{I}}$ (\textit{i.e.}, Dirac distribution) when $\mathcal{D} = \{\rm{I} \in \mathcal{X}^d\}$. Here ``Non-degenerate'' means $\vert \operatorname{supp}(\mathbb{P}_{{real}}) \rvert > 1$ and ``realistic'' satisfies $\operatorname{supp}(\mathbb{P}_{\theta})\subseteq \operatorname{supp}(\mathbb{P}_{real})$. 
Obviously, SIG is an ill-posed problem. In order to optimize it, appropriate regularization terms are needed to obtain the non-trivial solutions.
Thus, the objective function can be defined by:
 \begin{equation} \label{reg loss}
 \small
 	\min_{\theta} \rho(\delta_{\mathbf{I}},\mathbb{P}_{\theta}) + \varphi(\mathbb{P}_{\theta}),
 \end{equation}
where $\rho$ denotes the metric that measures the two distributions and $\varphi$ denotes the regularization term that contains prior knowledge to make $\mathbb{P}_{\theta}$ diverse and realistic.

\paragraph{Regularized latent variable model.} 
 
However, objective (\ref{reg loss}) is intractable due to the high dimensionality of $\mathbf{I}$ and the difficulty of correctly choosing $\varphi$. A more efficient way is to introduce a lower dimensional latent random variable $\boldsymbol{c} \in$ $\mathcal{C}$ to control $\boldsymbol{x}$.
Thus, the objective can be changed to learn the joint distribution $\mathbb{P}_{\theta}(\boldsymbol{c},\boldsymbol{x})=\mathbb{P}_{\theta_0}(\boldsymbol{c})\mathbb{P}_{\theta_1}(\boldsymbol{x}|\boldsymbol{c})$, which is easier to solve by considering $\mathbb{P}(\boldsymbol{c})$ and $\mathbb{P}(\boldsymbol{x}|\boldsymbol{c})$ respectively. Given the sample $\mathbf{I}$ and the corresponding latent code $\mathbf{c}_{\mathbf{I}}$, the optimization objective is:
\begin{equation} \label{latent loss}
\small
    \min_{\theta} \rho(\delta_{\mathbf{c}_{\mathbf{I}},\mathbf{I}},\mathbb{P}_{\theta}) + \varphi(\mathbb{P}_{\theta}),
\end{equation}
where $\delta_{\mathbf{c}_{\mathbf{I}},\mathbf{I}}$ is also a Dirac distribution of $(\mathbf{c}_{\mathbf{I}},\mathbf{I})$. 
Actually, all the previous SIG models including \cite{Shocher2019InGANCA,Shaham2019SinGANLA,Gur2020HierarchicalPV,Hinz2020ImprovedTF,Zheng_2021_CVPR,Zhang2021ExSinGANLA} can be applied by our formulation (\ref{latent loss}). Here, we take the canonical two-stage SinGAN \cite{Shaham2019SinGANLA} as an example to illustrate the definition and disadvantages of previous works.
As Fig.~\ref{fig:petsgan} shows, the first stage in SinGAN is to learn $\mathbb{P}_{G_0}(\boldsymbol{c})$, where $\boldsymbol{c}$ is downsampled image of $\boldsymbol{x}$. The learned distribution $\mathbb{P}_{G_0}$ determines the model capability in diversity. The second stage is to learn $\mathbb{P}_{G_1}(\boldsymbol{x}|\boldsymbol{c})$, which plays a role as super-resolution or restoration based on the generated latent code by $G_0$. The whole process can be formulated as:
\begin{small}
\begin{equation}\label{msloss}
\begin{split}
 	 &\min_{G_0, G_1} \ \rho_{\tau}(\delta_{\mathbf{c}_{\mathbf{I}}},\mathbb{P}_{G_0}) +
 	\rho_{\tau}(\delta_{\mathbf{I}},\mathbb{P}_{G_1}) +\varphi(\mathbb{P}_{G_0}) \\
 	&\textit{s.t.} \  \varphi(\mathbb{P}_{G_0})=\begin{cases}
 	 			0 & \text{if}\ G_0=\mathop{\arg\min}\limits_{G_0}\rho_{\tau}(\delta_{\mathbf{c}_{\mathbf{I}}},\mathbb{P}_{G_0})  \\ 
		\infty & \text{otherwise}
 	\end{cases}.
\end{split}
\end{equation}
\end{small}
The objective (\ref{msloss}) reveals the success of SinGAN: Firstly, the metric $\rho_{\tau}$ defined on \textit{patch distribution} is benefit for diverse layout and high-fidelity texture synthesis. Secondly, the regularization $\varphi$ suggests to train low-resolution generator $G_0$ and high-resolution generator $G_1$ progressively, which eases the instability of training. However, the drawbacks of SinGAN are also exposed: 
1) the first stage only focuses on learning the low-level information \textit{i.e.}, internal patches, lacking of the high-level semantic or structural information which is important to synthesize natural and plausible samples. 2) the cascaded learning not only takes a long time but also suffers from artifacts accumulation. These drawbacks make SinGAN perform poorly on the non-textured images, so do the derived models adopting the paradigm of SinGAN.


\subsection{PetsGAN} \label{PetsGAN}
\par To comprehensively tackle the SIG problem, we propose a Prior-based end-to-end training single image GAN (PetsGAN), which is composed of a Deep External Prior Network (DEPNet, denoted as $G$) and a Deep Internal Prior Network (DIPNet, denoted as $F$). The overall objective for PetsGAN is defined by:
\begin{equation}
\small
 	\label{our loss}
 	 	\min_{G,F}\  \rho_{\tau}(\delta_{\mathbf{I}},\mathbb{P}_{F\circ G}) + \varphi(\mathbb{P}_{G})  + \phi(\mathbb{P}_{F}), 
\end{equation}
where $\rho_{\tau}$ is the metric of patch distribution. $\varphi$ and $\phi$ are regularization terms corresponding to the external priors and the internal priors, respectively. 
As we have mentioned above, the patch distribution learning is a crucial factor for the success of SinGAN. Therefore, we adopt the same strategy to minimize the patch adversarial loss between the generated images and the original image. However, different from SinGAN which employs multiple discriminators for different stages, we only utilize one multi-scale discriminator (denoted as $D$) with different dilated kernel sizes without gradient penalty, so that the training time is reduced significantly. 
Moreover, we design novel regularization terms to achieve high-diversity in layout and high-fidelity in texture by fully utilizing the external priors and internal priors. Next, we will introduce the two parts in details.


\paragraph{External priors for high-diversity and reasonable layout.} 
The lack of high-level semantic and structural information makes SinGAN difficult to synthesize diverse and plausible samples, especially for non-texture images (\textit{e.g.,} object images). To alleviate this problem, we employ GAN inversion to inject external priors.
GAN inversion has been well studied \cite{zhu2016generative,pan2020exploiting,Huh2020TransformingAP} in recent years. Resorting to the pre-trained generator $G_{pre}$, image $\mathbf{I}$ can be projected into the latent space and obtained an optimal latent code $\boldsymbol{z}^*$, and then extensive samples 
can be synthesized by disturbing $\boldsymbol{z}^*$ with $\Delta \boldsymbol{z}$. 
Although the disturbance may reduce the quality of texture, it can generate abundant  syntheses covering a large variation in layout. We donwsample these syntheses to compress the texture details and utilize them to train a lightweight Deep External Prior Network (DEPNet). 
In practice, we define $\varphi(\mathbb{P}_{G})$ by
\begin{equation}  \label{external prior adv}
\small
\begin{split}
\varphi(\mathbb{P}_{G}) = & \mathbb{E}_{x\sim \mathbb{P}_{G_{pre}},z\sim \mathcal{N}(\mathbf{0},\mathbf{1} )}[\log{D_G(x)}-\log{D_G({G(z)})}] \\
& -\mathbb{E}_{z_1,z_2\sim  \mathcal{N}(\mathbf{0},\mathbf{1} )} \vert{D_{G}({G(z_{1})})-D_{G}({G(z_{2})}})\vert ,
\end{split}
\end{equation} 
where $G$ refers to DEPNet and $D_{G}$ is discriminator. 
The first term in Eq.(\ref{external prior adv}) is the non-saturating GAN loss \cite{goodfellow2014generative}. The second term is the collapse penalty to prevent $G$ from mode collapse. 
The noise is concatenated with the sinusoidal positional encoding as the input to enhance the positional bias of DEPNet \cite{Xu2020PositionalEA}. 

\paragraph{Internal priors for fast restoration of high-fidelity texture.} 
The remarkable trait of SIG models is that, it makes full use of patch discriminator \cite{Li2016PrecomputedRT,Isola2017ImagetoImageTW} to keep the synthesized texture consistent with that of the original image, \textit{i.e.}, minimizing $\rho_{\tau}(\delta_{\mathbf{I}},\mathbb{P}_{F})=\rho(\mathbb{P}^s_{\mathbf{I}},\mathbb{P}^s_{F})$. However, training patch discriminator is very tricky and time-consuming. Additional gradient regularization like gradient penalty \cite{gulrajani2017improved} or R1 regularization \cite{Mescheder2018WhichTM} is required to moderate the gradient from discriminator. Since the patch distribution $\mathbb{P}^s_{\boldsymbol{I}}$ is fixed, are there some 
efficient ways to approach it?  Here, we show that the trait can be theoretically realized effectively via the reconstruction loss. 
\begin{proposition}\label{proposition}
Considering image $\mathbf{I}$ and its downsampled version $\mathbf{c}_{\mathbf{I}}$, an image restoration network ($IR$) can be well designed such that $\mathbf{I} = {IR}(\mathbf{c}_{\mathbf{I}})$ and 
\begin{enumerate}[(i)]
    \item \label{proposition a} For appropriate $s$, there exist $s^* \leq s$ such that $IR$ determines a mapping from $\mathbb{P}^{s^*}_{\boldsymbol{\mathbf{c}_{\mathbf{I}}}}$ to $\mathbb{P}^{s}_{\mathbf{I}}$. 
    \item \label{proposition b} If  $\rho(\mathbb{P}^{s^*}_{\boldsymbol{\mathbf{c}_{\mathbf{I}}}}, \mathbb{P}^{s^*}_{\boldsymbol{c}})=0$, then
      $\rho(\mathbb{P}^s_{\mathbf{I}},\mathbb{P}^s_{{IR}(\boldsymbol{c})})=0.$
\end{enumerate}
\end{proposition}
\noindent Proof is provided in Appendix. Proposition  \ref{proposition} (\ref{proposition a}) exhibits that $IR$ trained by reconstruction loss not only  realizes pixel-level recovery, but also acts as a bridge between the patch distribution of low resolution images and that of high resolution images. (\ref{proposition b}) exposes an imaginative way to capture the trait of previous SIG models, \textit{i.e.},
reducing the differences of patch distribution between the generated samples and the original image in low resolution.
  
\par We inject internal priors of patch distribution correction into  DIPNet $F$ to realize fast restoration from low resolution syntheses to high resolution syntheses. 
DIPNet is a fully convolutional network containing two sub-modules, respectively patch transfer module ($PT$) for replacing each patch of synthesis $\boldsymbol{c}$ with the nearest patch of original image, and $IR$ module to restore the new synthesis. For the ${\boldsymbol{\omega}}^{\boldsymbol{c},s}_i$,  $PT$ learns the corresponding patch $a_{i,i^*} {\boldsymbol{\omega}}^{\mathbf{c_{\textbf{I}}},s}_{i^*}$, where $a_{i,i^*}$ (Eq. \eqref{attention}) is the coefficient from the attention map $\boldsymbol{A}=(a_{i,j})_{d\times d}$ to make the patch match differentiable.
\begin{small}
\begin{gather}
b_{i,j}=sim(Conv(\boldsymbol{\omega}^{\boldsymbol{c},s}_i),Conv(\boldsymbol{\omega}^{\mathbf{c_{\textbf{I}}},s}_j)\notag) \\
   (a_{i,j})_{j=1,\dots,d}=softmax(b_{i,1},\dots,b_{i,d}), \label{attention}  \\
    i^*= \mathop{\arg\max}\limits_j a_{i,j}. \notag
\end{gather}
\end{small}
The $sim$ denotes the cosine similarity, and $Conv$ is initialized by the pre-trained VGG \cite{Simonyan2015VeryDC}. The perceptual distance \cite{Johnson2016Perceptual} computed in feature space is more robust than in the pixel space \cite{zhang2019image}. These patches are integrated as a new synthesis, where overlap regions between different patches are averaged, then the new synthesis is fed into $IR$ to restore high resolution. In practice, we form $\phi(\mathbb{P}_F)$ as 
\begin{equation}
\small
\label{internal prior}  
    \phi(\mathbb{P}_F) = \lVert F(\mathbf{c}_{\mathbf{I}}+\Delta \mathbf{c})-\mathbf{I}\rVert-\lVert \boldsymbol{A} \rVert^2_F,
\end{equation}
where $\Delta \mathbf{c}$ is a small disturbance to make DIPNet more robust, and a sparsity constraint is imposed on $\boldsymbol{A}$ to make $a_{i,i^*}$ close to 1 (Remark 1 in Appendix). 
Fig. \ref{fig:patch_dis} shows the strength of internal priors. Although $IR$ can restore $\mathbf{I}$ from $\mathbf{c}_{\mathbf{I}}$ well (red images), for $ \boldsymbol{c} \sim \mathbb{P}_{G}$, $IR(\boldsymbol{c} )$ 
produces very poor texture (blue images). After patch transfer, $IR(PT(\boldsymbol{c}))$ has similar texture to $\boldsymbol{I}$ (cyan images).  
The phenomenon in Fig. \ref{fig:patch_dis} is in line with Proposition \ref{proposition}.
In Sec.\ref{sec:experiments} we also show its amazing ability to save training time on high resolution images. 

In summary, the PetsGAN minimize the patch generative adversarial loss $\rho_{\tau}$ and two regularization terms $\varphi$ and $\phi$ corresponding to external and internal priors. The joint training of DEPNet and DIPNet is conducive to mutual adaptation, and the ablation study also proves the effectiveness.

\begin{figure*}[t]
    \centering
    \includegraphics[scale=0.75]{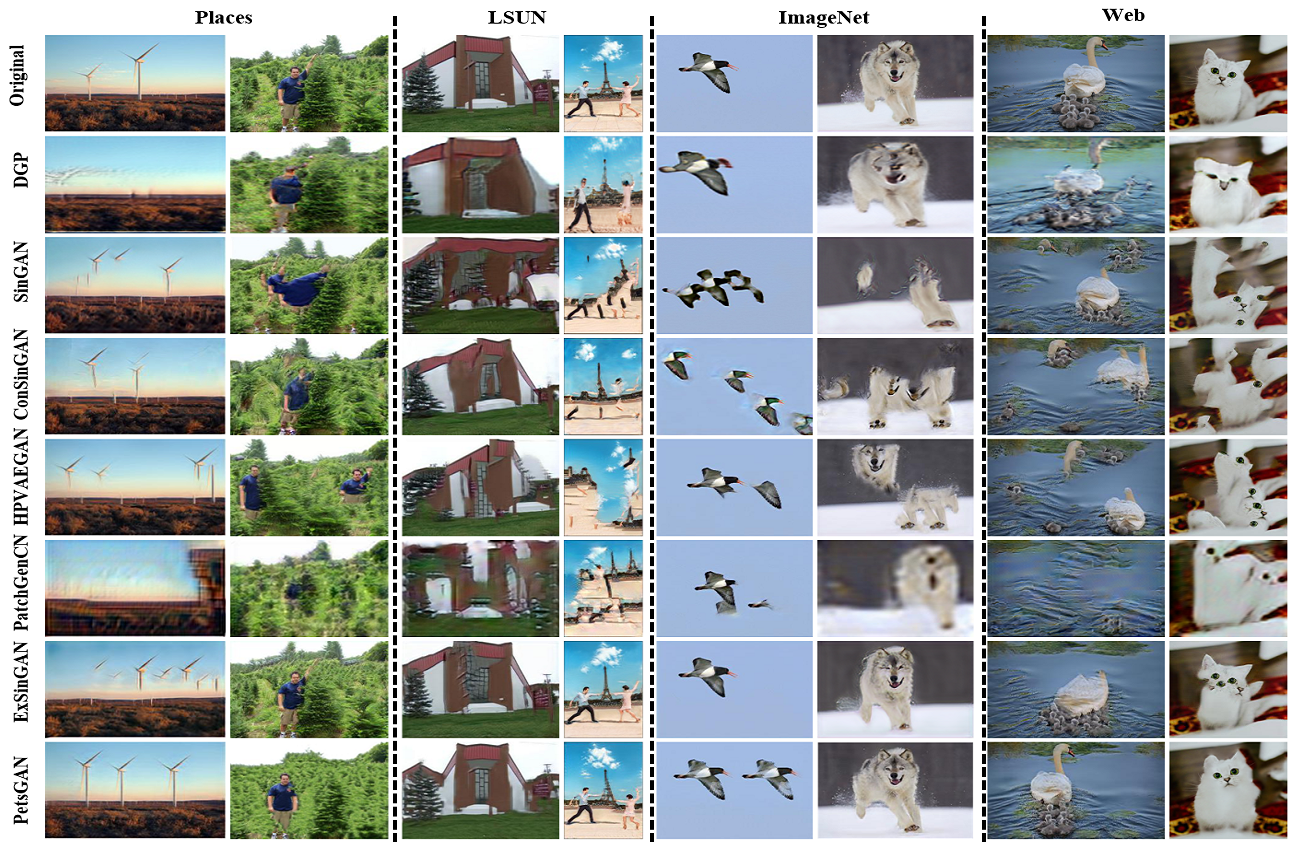}
    \caption{ Visualization results of different SIG models on different datasets. Our method has great advantage in the quality of
photo-realistic generation while guaranteeing the diversity, for not only scene but also object images.}
    \label{fig:qualitative comparison}
\end{figure*}

\section{Experiments}\label{sec:experiments}

In this section, we evaluate our method on four datasets and compare the results with other state-of-the-art SIG models, \textit{i.e.,} DGP \cite{pan2020exploiting}, SinGAN \cite{Shaham2019SinGANLA}, ConSinGAN \cite{Hinz2020ImprovedTF}, HPVAEGAN \cite{Gur2020HierarchicalPV}, PatchGenCN \cite{Zheng_2021_CVPR} ExSinGAN \cite{Zhang2021ExSinGANLA}, qualitatively and quantitatively.

\subsection{Training Settings}
\paragraph{Datasets.} 
For comprehensively evaluating the generative capability of PetsGAN, various experimental images are collected from various sources. The first one is the Places50 dataset \cite{Shaham2019SinGANLA}, which contains 50 natural landscape scene images, the second one is the LSUN50 dataset \cite{Hinz2020ImprovedTF}, which contains 50 images with more complex scene images, and the third one is the ImageNet50 dataset \cite{Zhang2021ExSinGANLA}, which contains 50 images of objects. Besides, we also qualitatively show syntheses of images from the internet to illustrate the generalization of PetsGAN.

\paragraph{Training Details.}
Our experiments are conducted on NVIDIA RTX 3090. The input image $\mathbf{I}$ is resized to make the longer side no more than $256$, then downsampled with scale factor $8$ to obtain the low-resolution image $\mathbf{c}_{\mathbf{I}}$.  We adopt DGP \cite{pan2020exploiting} as the inversion method, in which disturbance is sampled from $\mathcal{N}(0,0.5)$. The input noise of PetsGAN is sampled from $\mathcal{N}(0,1)$. The window size $s$ is set to 7. We use Adam optimizer for  $G, F, D_{G}, D$ but with different learning rates. The training epoch of PetsGAN is $5000$. The batch sizes for optimizing $\rho_{\tau}(\delta_{\mathbf{I}},\mathbb{P}_{F\circ G})$ and $\varphi({\mathbb{P}_{G}})$ are $1$ and $32$, respectively. 
The warm up strategy is used to train $G$ and $F$. Due to space limitation, the detailed network structures and learning strategy are provided in the Appendix.

\subsection{Experimental Results}
\paragraph{Qualitative analysis.}
Fig.~\ref{fig:show} gives syntheses of PetsGAN on various exemplars  with arbitrary sizes. We can see PetsGAN performs excellent on both texture and object images. \textbf{It can generate images with diverse layouts while the content and texture are consistent with the exemplars}.
To demonstrate the superiority of PetsGAN, we compare PetsGAN with other SIG models. As Fig. \ref{fig:qualitative comparison} shows, SinGAN and ConSinGAN usually generate unnatural layouts or structures due to lack of semantic information, \textit{e.g.,} windmills and human bodies. DGP utilizes the external information to generate reasonable structures but the texture is synthesized coarsely. ExSinGAN also embodies the issue of diversity because the semantic consistence constraint between the synthesized samples and the source image. By contrast, PetsGAN surpasses them in terms of both diversity and quality.

\begin{figure}[t]
    \centering
    \includegraphics[scale=0.3]{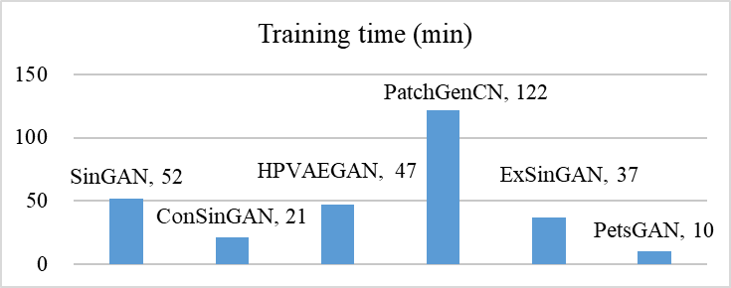}
    \caption{Training time of different SIG models.}
    \label{fig:training time}
\end{figure}

\paragraph{Quantitative analysis.}
We compare the performance of SIG models in terms of generative quality, diversity, and training time quantitatively. 
Similar to ExSinGAN, we adopt SIFID and LPIPS to measure the image quality and diversity, respectively. SIFID quantifies the distance between the generated image and the original image in feature space. LPIPS \cite{zhang2018unreasonable} compares the perceptual differences using  pre-trained AlexNet, which is insensitive to color changes but more sensitive to spatial changes. The measured values are provided in Tab.~\ref{table1}. We find that the similar conclusion can be drawn as the qualitative analysis. 
Although some methods achieve high scores in LPIPS, they perform poorly in SIFID, and the visualization results are bad in this case.
Compared with other methods, PetsGAN achieves low scores in SIFID and high scores in LPIPS simultaneously, which means PetsGAN can generate diverse samples with natural layout and realistic texture.
Besides, our model has significant advantages in training efficiency. As Fig.~\ref{fig:training time} shows, the average training time of our model is 10 minutes, almost 4 times faster than SinGAN.

\paragraph{Ablation study.}
To better evaluate the effect of  training strategies and priors, we conduct a series of ablation studies (Table \ref{table1}).
Firstly, comparing the joint training with cascaded training, the former's SIFID is consistently smaller than the latter, conditioned on similar LIPIS, proving
the joint training indeed helps DIPNet and DEPNet towards better mutual adaption. 
Secondly, patch adversarial loss is taken out, namely training DEPNet and DIPNet independently, and thereby the SIFID increases largely, proving its positive contributions to synthesis. Thirdly, if we do not use external priors, with plummeting LPIPS the syntheses lose the diversity largely, meaning that external prior is the key to provide the diversity, and avoid the overfitting. Finally, the SIFID and LPIPS both become worse if we do not use internal prior. This implies that the internal prior is not only helpful to the quality, but also prevents over-fitting to some extent. In summary, the ablation study demonstrated the rationality of PetsGAN's design and the necessity of each part.

\begin{table}[t]
    \centering
    \setlength\tabcolsep{0.2pt}
    {
        \begin{tabular}{@{}c|c|c|c|c|c|c@{}}
        	\hline\hline
				  & \multicolumn{2}{c|}{{{Places50}}} & \multicolumn{2}{c|}{{{LSUN50}}} & \multicolumn{2}{c}{{{ImageNet50}}} \\
				\cline{2-7}
				& SIFID$\downarrow$   & LPIPS$\uparrow$    & SIFID$\downarrow$   & LPIPS$\uparrow$   & SIFID$\downarrow$     & LPIPS$\uparrow$ \\
				\hline\hline
					DGP  & 0.64 & 0.28 &0.75 &0.36 &1.65&0.31 \\ 
				SinGAN     & 0.09    & 0.25   & 0.23    & 0.33    & 0.60     & 0.33   \\
				
				ConSinGAN              & {0.06}    & 0.24    & 0.11    & 0.32   & 0.56     & 0.39       \\
				
			    HPVAEGAN &0.30 &0.38  &0.35 &0.44 & 0.55 &0.47\\ 
			    PatchGenCN &1.23 &0.60& 1.24&0.62 &1.29 &0.61 \\
				ExSinGAN               & 0.10    & {0.23}   & {0.11}    & {0.25}   & {0.45}     & {0.24}   \\
			    \cline{1-7}
				PetsGAN               & 0.08    & {0.31}     & {{0.09}}    & {0.33}     & {{0.21} }    & 0.35  \\
				\hline\hline
		cascaded & 0.10    & {0.31}     &{0.11}    & {0.34}    &{0.50}     & 0.37  \\
		w/o $\rho_{\tau}(\delta_{\mathbf{I}},\mathbb{P}_{F\circ G})$ &   0.23  & 0.33     & {0.23}    & {0.38}     & {0.30}     & 0.39  \\
		w/o $\varphi(\mathbb{P}_{G})$& 0.11 &0.18 &0.11 &0.21 &0.25 &0.22 \\
		w/o $\phi(\mathbb{P}_{F})$& 0.20 &0.23  &0.19 &0.26 &0.40 &0.33  \\ 		
		\hline\hline		
			
		\end{tabular}
    }
		
		\caption{Quantitative results of different SIG models and ablation study.
		``cascaded'' means training DEPNet individually, and then keeping it fixed and training DIPNet.
		}
		\label{table1}
	\end{table}

\begin{figure}[t]
    \centering
    \includegraphics[scale=0.39]{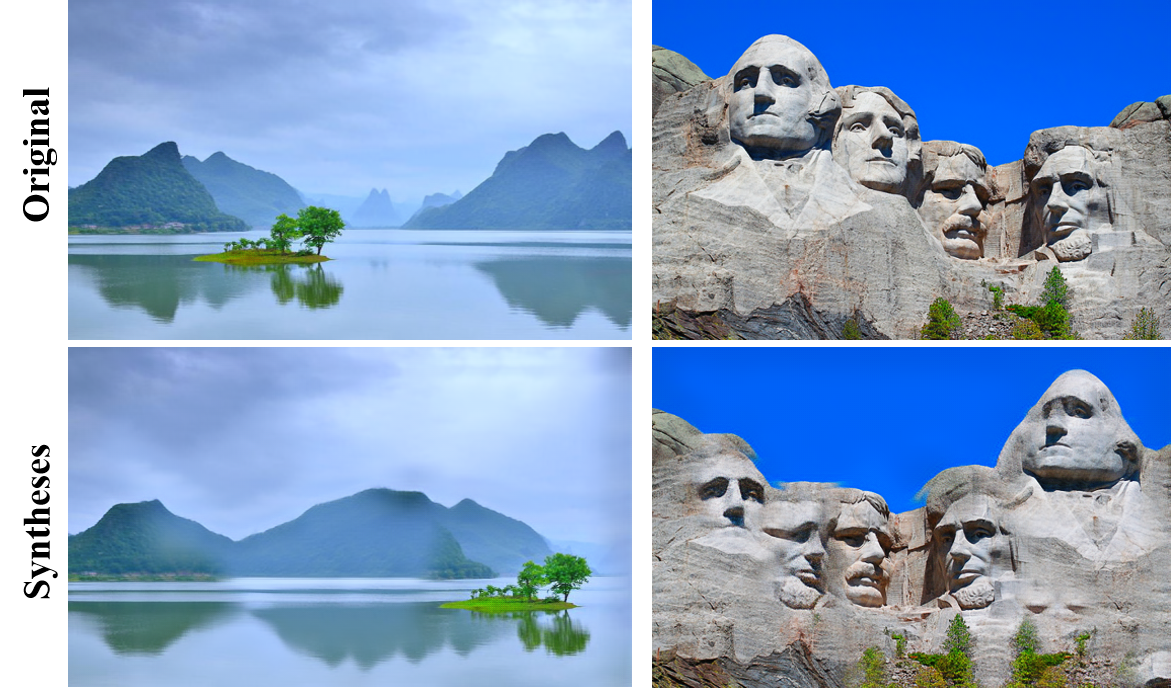}
    \caption{High resolution image generation.}
    \label{fig:applications1}
\end{figure}

\begin{figure}[t]
    \centering
    \includegraphics[scale=0.4]{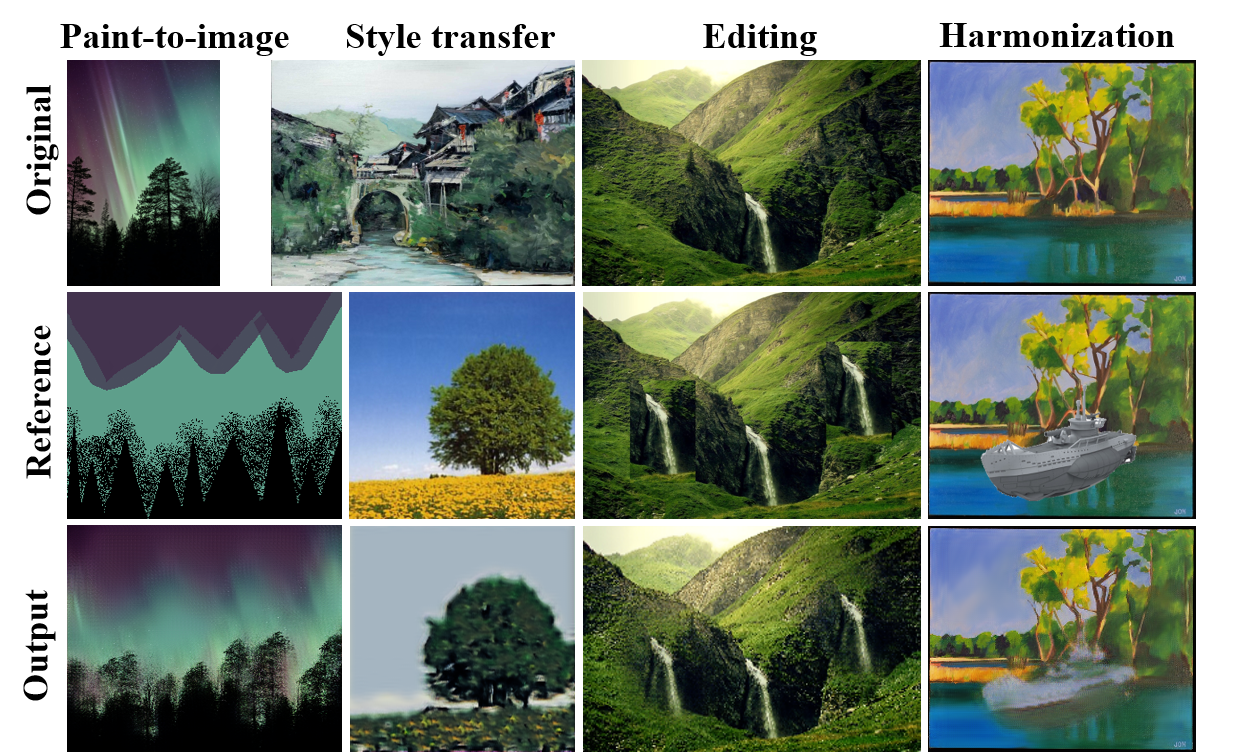}
    \caption{Applications to other image manipulation tasks.}
    \label{fig:applications2}
    \vspace{-0.1cm}
\end{figure}
\subsection{More Applications} \label{sec:alpplications}

\paragraph{Fast training of high resolution images.}
PetsGAN can also handle high resolution image generation by simple plugging in a new DIPNet, while keeping high efficiency in training.
SinGAN takes about 6 hours for training with an image of size  $1024\times1024$ on NVIDIA RTX 3090. The time-consuming training undoubtedly hinders the application prospect of SIG models.  By contrast, we first obtain the $256\times256$ synthesis, and feed it into the new DIPNet with 4 times upscaling only trained by reconstruction loss.  Thus, the overall training can be completed in less than 15 minutes, and the synthesized texture is of high-fidelity without degradation (Fig. \ref{fig:applications1}) benefiting from the internal prior. 
\paragraph{Image manipulation tasks.} 
Our model can also be applied to many other image manipulation tasks, \textit{e.g.}, paint-to-image, style transfer, editing and harmonization. A recent work SinIR \cite{Yoo2021SinIREG} shows that the pyramid networks only with reconstruction loss can do these tasks effectively. Different from SinIR, we only use the one stage DIPNet with reconstruction loss to realize these applications. We show some results of our method in Fig.~\ref{fig:applications2}. The training time cost is less than 2 minutes,  which is much faster than SinIR and other SIG models, and the generated results are also satisfactory.
\section{Conclusion}
In this paper, we focus on solving the SIG problem by fully-utilizing the prior knowledge. First, we formulate SIG with a regularized latent model, and analyze the shortcomings of SinGAN and other work from the form. Based on latent model, we propose PetsGAN using external and external priors. The external priors are used to provide high-level information and prevent model overfitting. The internal prior is to reduce the patch distribution discrepancy between the syntheses and the input image, which can effectively speed up the training of the model. PetsGAN can achieve competitive results and improvements for SIG with more image types and faster training speed. It can also be applied to image manipulate tasks at a high speed.
\section*{Acknowledgements} This paper is supported by the National key research and development program of China (2021YFA1000403), the National Natural Science Foundation of China (Nos. U19B2040, 11731013, 11991022), the Strategic Priority Research Program of Chinese Academy of Sciences, Grant No. XDA27000000, and the Fundamental Research Funds for the Central Universities.
\bibliography{aaai22}
\end{document}